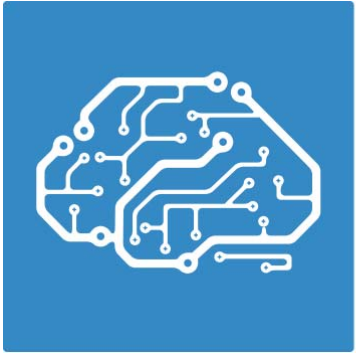

CBMM Memo No. 21                                      June 14, 2014

# Neural tuning size is a key factor underlying holistic face processing

by

Cheston Tan and Tomaso Poggio

Abstract: Faces are a class of visual stimuli with unique significance, for a variety of reasons. They are ubiquitous throughout the course of a person's life, and face recognition is crucial for daily social interaction. Faces are also unlike any other stimulus class in terms of certain physical stimulus characteristics. Furthermore, faces have been empirically found to elicit certain characteristic behavioral phenomena, which are widely held to be evidence of "holistic" processing of faces. However, little is known about the neural mechanisms underlying such holistic face processing. In other words, for the processing of faces by the primate visual system, the input and output characteristics are relatively well known, but the internal neural computations are not. The main aim of this work is to further the fundamental understanding of what causes the visual processing of faces to be different from that of objects. In this computational modeling work, we show that a single factor – "neural tuning size" – is able to account for three key phenomena that are characteristic of face processing, namely the Composite Face Effect (CFE), Face Inversion Effect (FIE) and Whole-Part Effect (WPE). Our computational proof-of-principle provides specific neural tuning properties that correspond to the poorly-understood notion of holistic face processing, and connects these neural properties to psychophysical behavior. Overall, our work provides a unified and parsimonious theoretical account for the disparate empirical data on face-specific processing, deepening the fundamental understanding of face processing.

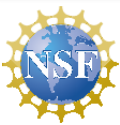 This work was supported by the Center for Brains, Minds and Machines (CBMM), funded by NSF STC award CCF-1231216.

# Neural tuning size is a key factor underlying holistic face processing

**Authors:** Cheston Tan[a,b,c], Tomaso Poggio[a,b]

**Author affiliations:**

[a] McGovern Institute for Brain Research, MIT, Cambridge, MA 02139, USA.

[b] Department of Brain and Cognitive Sciences, MIT, Cambridge, MA 02139, USA.

[c] Institute for Infocomm Research, Singapore 138632, Singapore.

**Corresponding author:**

Name: Cheston Tan

Address: 1 Fusionopolis Way, #21-01 Connexis (South Tower), Singapore 138632.

Phone: +65 6408 2561

E-mail: cheston-tan@i2r.a-star.edu.sg

Note: this technical report supersedes and replaces a previous report (40): MIT CSAIL Technical Report 2013-004 / CBCL Memo 311. URI: http://hdl.handle.net/1721.1/77936

Introduction

Faces are an important class of visual stimuli with unique significance, and face processing is a longstanding topic of active study within neuroscience (1, 2), cognitive psychology (3-5) and artificial intelligence (6-8). Faces are ubiquitous throughout the course of a person's life, and face recognition is important for daily social interaction. Faces are also unlike any other stimulus class in terms of certain defining physical stimulus characteristics, such as having a very specific configuration of internal parts. Furthermore, faces have been empirically found to elicit certain characteristic behavioral phenomena (e.g. the Composite Face Effect), which are widely held to be evidence of "holistic" processing of faces. However, little is known about the neural mechanisms underlying such holistic face processing. In other words, for the processing of faces by the primate visual system, the input and output characteristics are relatively well known (or are straightforward to measure), as are the neural correlates (1, 4, 9), but the actual neural computations are not known.

Here, we investigate the nature of the difference between face and (non-face) object processing at the level of the underlying computational mechanisms, in order to further the fundamental understanding of what causes the visual processing of faces to be different from that of objects. In this computational modeling work, we show that a single factor – "neural tuning size" – is able to account for three key phenomena that are characteristic of face processing, namely the Composite Face Effect (CFE), Face Inversion Effect (FIE) and Whole-Part Effect (WPE). Our computational proof-of-principle provides specific neural tuning properties that correspond to the poorly-understood notion of "holistic" face processing, and shows the connection between these neural properties and empirical psychophysical behavior. Our work predicts that such tuning properties will be found only in neurons associated with holistic face processing, e.g. in the macaque Fusiform Face Area (FFA) but not the Occipital Face Area (OFA). Overall, our results provide a unified and parsimonious theoretical account for the disparate empirical data on face-specific processing.

Results

Building upon the family of simple, biologically-plausible visual recognition models (10-14), we found that a single parameter determines whether processing is "face-like" or "object-like", as gauged by three important face-specific phenomena. The first, the Composite Face Effect (CFE) (15), is the phenomenon whereby two identical top halves are sometimes incorrectly perceived as different, when paired with different bottom halves (Fig. 1a). This effect is ostensibly due to the top and bottom halves of each composite being perceived "holistically" (together as a whole) when aligned, despite instructions to ignore the bottom halves. Perception is more accurate when the halves are misaligned (Fig. 1b). Crucially, this effect occurs only for faces, and is therefore commonly taken as evidence that face and object processing are qualitatively different (16-19).

We probed the minimal conditions required to produce – or abolish – such holistic face processing, and found that a vital factor is the size of the template that specifies the tuning profile and optimal stimulus of each neuron (henceforth termed "tuning size"). Tuning size is defined in terms of proportion of a whole face covered by a template (see Methods section), not in terms of number of pixels or degrees of visual angle. When tuning size is large, even without encompassing the whole face, the Composite Face Effect is found (Fig. 1c). A single change – reduction of tuning size – abolishes the Composite Face Effect, i.e. leads to "object-like" processing of faces.

"Holism" is a psychological construct with multiple interpretations and putative mechanisms (19, 21, 22) for which a consensus has yet to emerge. Our simulation results argue in favor of one particular interpretation of holism, that it is simply the byproduct of having large tuning size (18, 23, 24). The Composite Face Effect is found using each individual model neuron with large tuning size by itself (Fig. 1c inset), even though tuning size is less than half the whole face, and there is no overlap between receptive fields to speak of. Conversely, even though the set of neurons with small tuning size collectively span the whole face, they do not produce the CFE (Fig. 1c).

Since there is nothing qualitatively "whole", "singular", "unified", "global" or "non-decomposable" about processing that uses large tuning size rather than small, the term "holistic" may be somewhat misleading (to the extent that it implies a qualitative difference, an absolute whole, or integration into a single representation). Our results do, however, indicate that holism is not an all-or-none phenomenon, but it is one that can vary continuously depending on tuning size (and be modulated by other factors).

In our simulations, tuning size is the sole change between the two conditions depicted in Fig. 1c (large and small tuning size), which suggests that decisional and attentional factors are not key. Rather, what matters is the amount of "perceptual integration", as controlled by tuning size. Additionally, while detection and segmentation are important processes for accurate face recognition, the absence of explicit mechanisms for these in our simulations suggest that they are also not key factors relating to holism.

Tuning size also accounts for another key face-specific phenomenon, the Face Inversion Effect (FIE), whereby upside-down inversion disrupts face processing significantly more than object processing (25). We found that when tuning size is reduced, the behavioral effect of inversion is also reduced (Fig. 2), akin to face processing becoming "object-like". Our simulations also show that inversion reduces the mean response of each individual neuron (Fig. 3a), illustrating the neural basis of the behavioral Face Inversion Effect (26). More importantly, tuning size also controls the magnitude of the neural FIE (Fig. 3b). Even after taking into account the coverage areas corresponding to different tuning sizes (Figs. 2e, f) or taking into account the mean response in the upright condition (Figs. 3c, d), large tuning size still produces significantly larger behavioral and neural FIEs than small tuning size (see Methods section for details).

Finally, tuning size also accounts for the Whole-Part Effect (27), a "gold-standard" test of holistic processing (16), like the CFE. In the Whole-Part Effect (WPE), subjects are first presented a study face (Fig. 4a) to memorize. At test, in the Whole condition, subjects are presented with two faces (Fig. 4b) that differ only in a localized region (e.g. eye region), and have to recall and choose the study face. In the Part condition, the differing localized regions in the Whole condition are cropped (Fig. 4c) and presented instead. Empirically, subjects are significantly more accurate in the Whole than Part condition. This effect is highly significant for faces, but non-significant or significantly smaller for non-faces (16). Our simulations show that reduction in tuning size results in a significantly smaller WPE (Fig. 4e), akin to a shift from "face-like" to more "object-like" processing.

Discussion

Clearly, our simulations do not capture the full complexity of face processing, including aspects such as gaze, expression and attentional capture. Also, tuning size is not necessarily the only difference between face and object processing mechanisms (28-30). However, we have shown that a change in tuning size alone can account for three important phenomena commonly thought to be characteristic of face processing, suggesting that tuning size is a key factor underlying holism.

By changing only tuning size and keeping everything else unchanged (Figs. 1-4), our simulations are able to sidestep a confound that is unavoidable for empirical experiments that investigate mechanisms underlying face versus object processing – the confound of different stimuli. Empirically, face and object stimuli elicit measurable differences, but do these stem from differences in physical stimulus properties, or from differences in processing mechanisms? Here, instead of changing the stimuli to produce measurable differences, we changed only the underlying processing but not the stimuli (we used only faces throughout), so this confound is avoided.

Our results suggest that all three effects stem from a common cause: large tuning size. This makes intuitive sense for a variety of reasons. Large tuning size is consistent with the existence of face-selective neurons being tuned to multiple face parts, but not necessarily the entire face (31, 32). Face recognition algorithms that use features with such characteristics (large but not necessarily whole-face tuning size) show excellent performance (6). While face processing in general has a significant innate component (33-35), visual deprivation during infancy abolishes the Composite Face Effect (36). This suggests that large tuning size may result from a combination of innate factors and postnatal visual experience.

Other than large tuning size, another interpretation of holistic processing is that of "non-linear interaction of parts" (18, 37). Since our model is highly non-linear, our results and this interpretation are not mutually exclusive. Our results do not preclude certain non-linearities from also being key factors underlying holistic processing, nor do they deny that non-linear interactions occur empirically.

Apart from "holistic" processing, faces are also thought to elicit "configural" processing, but the relationship between these notions is unclear (16, 19, 38). For instance, the Face Inversion Effect has sometimes been associated with configural rather than holistic processing of faces. Our results suggest a link between these two concepts, through the common factor of large tuning size. Because neurons with large tuning size cover more than individual face parts, they are more sensitive to the configuration of multiple parts, which is altered by inversion. For this same reason, these neurons are also more sensitive to information that comes from more distant regions of the face image (in the case of the Composite Face Effect and Whole-Part Effect). The notion of large tuning size may also be able to account for sensitivity to spacing between face parts. The idea is that since each neuron's tuning is specified by a certain face template, any deviation from that template, such as changing the distance between the eyes, will reduce the neural response. Note, however, that these neurons are not necessarily more sensitive to spacing changes than shape changes, consistent with (39).

Methods

**Model.** The HMAX model (10) simulates hierarchical processing in primate visual cortex, reflecting the increase in neural tuning complexity and invariance up the hierarchy. The lowest levels correspond to orientation-selective cells in primary visual cortex, while the highest levels correspond to face-selective and object-selective cells in inferotemporal cortex.

We used the model implementation found at http://cbcl.mit.edu/jmutch/cns/. Of the four model layers, the orientation-selective lower two layers (*S1* and *C1*) contain model neurons tuned to Gabor patches of various orientations and spatial frequencies; the parameters have been pre-determined based on prior electrophysiological data. The upper two layers (*S2* and *C2*) contain model neurons that are tuned during an unsupervised template-learning process, performed prior to normal model operation. Template-learning simply involves storing "snapshots" of *C1* activity produced in response to some set of training images. Subsequently, in normal model operation, these snapshots act as templates that new images are matched against. The *S2* layer comprises the output of this template-matching process, and pooling of the *S2* model neurons over all image positions and scales (to produce invariance to these) produces the *C2* layer. If training images consist of faces, then *S2* and *C2* model neurons are face-selective. All simulations used 1000 *C2* model neurons.

**Tuning size.** The critical independent variable is "tuning size". Large, medium and small tuning sizes correspond respectively to *S2* tuning templates covering 12x12, 8x8 and 4x4 *C1* model neurons, all from the relatively coarse scale 7 (out of 9). At this scale, the entire face oval corresponds to 17x22 *C1* neurons, so each small template is roughly the size of a face part (e.g. eye, nose), while <u>each large template covers multiple face parts but not the whole face</u>.

Importantly, "tuning size" is defined as the proportion of a whole face covered by a template. This is not the same as "size" defined in terms of number of pixels or degrees of visual angle. In the human and primate visual systems (as well as our model), there exists some invariance to image scale. Therefore, a particular tuning size (e.g. half a face) can correspond to a range of physical sizes (in pixels or degrees of visual angle).

Since each template was learnt from a different (random) part of a training image, even the 1000 small templates (collectively) spanned the whole face – yet they did not produce a Composite Face Effect (Fig. 1c), thus ruling out some alternative accounts of mechanisms underlying "holistic processing".

**Stimuli.** Stimuli were derived from 100 frontal-view male faces from the MPI database (http://faces.kyb.tuebingen.mpg.de/). Faces were downscaled by 25%, and then oval-cropped to remove outline and external features (e.g. hair). Faces were normalized so that all had the same pixel-value statistics (mean and variance). Odd-numbered faces were used for template-learning, even-numbered faces for normal operation. All faces were upright unless explicitly inverted. Note: in Figs. 1, 2 and 4, backgrounds were cropped slightly to save space and make face details more apparent.

**Composite Face Effect (CFE).** Composites were constructed by pairing the top of one face with the bottom of a different face (with a two-pixel gap). Twenty faces were used; these were chosen prior to all simulations, for behavioral replication of the CFE in human subjects (not related to this modeling study).

On each trial, two composites are presented, and their top halves are judged to be same or different, ignoring the bottom halves. Only trials with identical top halves are analyzed (16). The Composite Face Effect is defined as a higher hit-rate (i.e. accuracy on these *"same"* trials) for misaligned than aligned composites.

To simulate human subjects looking and attending to the top halves, bottom-half pixel values are multiplied by 0.1, and faces shifted downwards so that the top halves occupy the center. To simulate subjects comparing composites, if the dissimilarity between composites (Euclidean distance between the two sets of *C2* layer responses) is below some threshold, the composites are considered *"same"*. For each condition (e.g.

small tuning size), the threshold is independently set so that the aligned, upright hit-rate is as close to 75% as possible, but results are qualitatively robust to the threshold used.

Fig. 1c: p-values and error bars computed using 1,000 runs of non-parametric bootstrap (N=20 pairs of composites). Fig. 1c inset: p-values computed using non-parametric Wilcoxon signed-rank test (N=1000 independent $C2$ features).

**Face Inversion Effect.** Dissimilarity between two images was defined as the Euclidean distance between the two sets of $C2$ layer responses. Fig. 2c shows the mean dissimilarity between all 1225 (i.e. $^{50}C_2$) pairs of faces within each condition, averaged over bootstrap runs. Fig. 2d shows the mean behavioral FIE effect size (i.e. upright – inverted dissimilarities shown in Fig. 2c) over bootstrap runs. Fig. 2c, d: p-values and error bars computed using bootstrap runs. In all cases for Fig. 2, 10,000 runs of non-parametric bootstrap were used.

Figs. 2e and 2f are similar to 2c and 2d, except that for the large and medium tuning sizes, a random subset of neurons (100 for large, 150 for medium) were used in each bootstrap run, instead of all 1000 neurons. This was to control for the larger coverage areas (12x12 vs. 8x8 vs. 4x4 $C1$ neurons for large, medium and small respectively, and also to control for the different upright dissimilarities (see black bars in Fig. 2c); both factors are actually over-compensated for.

Fig. 3a shows the mean response (averaged over all 1000 model neurons) to all 50 faces. Fig. 3b shows the mean neural FIE (i.e. upright – inverted responses shown in Fig. 3a). Fig. 3a, b: p-values computed using non-parametric Wilcoxon signed-rank test.

Figs. 3c and 3d were similar to Figs. 3a and 3b, except that only neurons with a mean response between 0.75 and 0.80 were considered (to avoid ceiling and floor effects), after which neurons were then randomly chosen so that all conditions (i.e. tuning sizes) ultimately used the same number of neurons. This was to control for the slightly different upright responses in Fig. 3a (black bars).

**Whole-Part Effect (WPE).** Stimuli were constructed by blending the eye region of one face with the rest of a different face. Stimuli in the Part condition (Fig. 4c) were produced by cropping out the eye region of stimuli in the Whole condition (Fig. 4b). All possible trials (3-way combinations of first eye region, second eye region, rest of face) for twenty original faces (same faces as for CFE) were tested.

The Whole-Part Effect is defined as a higher accuracy for the 2-alternative-forced-choice task in the Whole condition than in the Part condition. To simulate human subjects looking and attending at the eye regions of the test faces, non-eye-region pixel values are multiplied by 0.5, and faces shifted downwards so that the eye regions occupy the center. To simulate subjects choosing between the two test stimuli, the one with the smaller dissimilarity (Euclidean distance between the *C2* layer responses) to the study stimulus (Fig. 4a) was picked. Fig. 4d, e: p-values and error bars computed using 1,000 runs of non-parametric bootstrap.

Acknowledgements

We thank G. Cottrell, R. Desimone, J. DiCarlo, I. Gauthier, N. Kanwisher, J. Leibo, D. Sheinberg, P. Sinha and the Perceptual Expertise Network for helpful discussions. This material is based upon work supported by the Center for Brains, Minds and Machines (CBMM), funded by NSF STC award CCF-1231216. This work was also supported by A*STAR Joint Council Office (JCO) Visiting Investigatorship Programme (VIP) grant #1335h00098. This work was also supported by DARPA (IPTO and DSO), National Science Foundation (NSF-0640097, NSF-0827427), AFSOR-THRL (FA8650-05-C-7262). Additional support (T. P.) was provided by: Adobe, Honda Research Institute USA, KAUST, NEC, Sony and the Eugene McDermott Foundation. C. T. was supported by fellowships from A*STAR and the McGovern Institute for Brain Research.

# **References**

1. Tsao DY, Freiwald WA, Tootell RBH, Livingstone MS (2006) A cortical region consisting entirely of face-selective cells. *Science* 311(5761):670-674.

2. Desimone R, Albright TD, Gross, CG, Bruce C (1984) Stimulus-selective properties of inferior temporal neurons in the macaque. *J Neurosci* 4(8):2051-2062.

3. Rossion B, et al. (2000) The N170 occipito-temporal component is delayed and enhanced to inverted faces but not to inverted objects: an electrophysiological account of face-specific processes in the human brain. *Neuroreport* 11(1):69-74.

4. Haxby JV, et al. (1996) Face encoding and recognition in the human brain. *Proc Natl Acad Sci USA* 93(2):922-927.

5. Pitcher D, Walsh V, Yovel G, Duchaine B (2007) TMS evidence for the involvement of the right occipital face area in early face processing. *Curr Biol* 17(18):1568-1573.

6. Viola P, Jones MJ (2004) Robust real-time face detection. *Int J Comput Vision* 57(2):137-154.

7. Brunelli R, Poggio T (1993) Face recognition: features versus templates. *IEEE Trans. Pattern Anal. Mach. Intell.* 15(10):1042-1052.

8. Turk M, Pentland A (1991) Eigenfaces for recognition. *J Cogn Neurosci* 3(1):71-86.

9. Kanwisher N, McDermott J, Chun MM (1997) The fusiform face area: a module in human extrastriate cortex specialized for face perception. *J Neurosci* 17(11):4302-4311.

10. Riesenhuber M, Poggio T (1999) Hierarchical models of object recognition in cortex. *Nature Neurosci* 2(11):1019-1025.

11. Cottrell GW, Branson KM, Calder AJ (2002) in *Proceedings of the 24th Annual Conference of the Cognitive Science Society*, (Cognitive Science Society, Mahwah, NJ) pp 238-243.

12. Fukushima K (1980) Neocognitron: A self-organizing neural network model for a mechanism of pattern recognition unaffected by shift in position. *Biol Cybern* 36(4):193-202.

13. Perrett DI, Oram MW (1993) Neurophysiology of shape processing. *Image Vision Comput* 11(6):317-333.

14. Wallis G, Rolls ET (1997) Invariant face and object recognition in the visual system. *Prog Neurobiol* 51(2):167-194.

15. Young AW, Hellawell D, Hay DC (1987) Configurational information in face perception. *Perception* 16(6):747-759.

16. McKone E, Robbins R (2011) in *The Oxford Handbook of Face Perception,* eds Calder AJ, Rhodes G, Johnson MH, Haxby JV (Oxford Univ. Press, New York, NY), pp 149-176.

17. Tanaka JW, Gordon I (2011) in *The Oxford Handbook of Face Perception,* eds Calder AJ, Rhodes G, Johnson MH, Haxby JV (Oxford Univ. Press, New York, NY) pp 177-194.

18. Tsao DY, Livingstone MS (2008) Mechanisms of face perception. *Annu Rev Neurosci* 31:411-437.

19. Richler JJ, Palmeri TJ, Gauthier I (2012) Meanings, mechanisms, and measures of holistic processing. *Front Psychol*, 3:553.

20. McKone E, et al. (2013) Importance of the inverted control in measuring holistic face processing with the composite effect and part-whole effect. *Front Psychol* 4:33.

21. Maurer D, Le Grand R, Mondloch CJ (2002) The many faces of configural processing. *Trends Cogn Sci* 6(6):255-260.

22. Piepers DW, Robbins RA (2012) A review and clarification of the terms "holistic", "configural" and "relational" in the face perception literature. *Front Psychol*, 3:559.

23. Rossion B, Gauthier I (2002) How does the brain process upright and inverted faces? *Behav Cogn Neurosci Rev* 1(1):63-75.

24. Farah MJ, Wilson KD, Drain M, Tanaka JN (1998) What is "special" about face perception? *Psychol Rev* 105(3):482-498.

25. Yin RK (1969) Looking at upside-down faces. *J Exp Psychol* 81(1):141-145.

26. Yovel G, Kanwisher N (2005) The neural basis of the behavioral face-inversion effect. *Curr Biol* 15(24):2256-2262.

27. Tanaka JW, Farah MJ (1993) Parts and wholes in face recognition. *Quart J Exp Psych* 46A:225-245.

28. Wallis G (2013) Toward a unified model of face and object recognition in the human visual system. *Front Psychol* 4:497.

29. Riesenhuber M, Wolff BS (2009) Task effect, performance levels, features, configurations, and holistic face processing: a reply to Rossion. *Acta Psychol* 132(3):286-292.

30. Wallraven C, Schwaninger A, Bulthoff HH (2005) Learning from humans: computational modeling of face recognition. *Network: Comp Neural Sys* 16:401-418

31. Freiwald WA, Tsao DY, Livingstone MS (2009) A face feature space in the macaque temporal lobe. *Nature Neurosci* 12(9):1187-1196.

32. Perrett DI, Rolls ET, Caan W (1982) Visual neurones responsive to faces in the monkey temporal cortex. *Exp Brain Res* 47(3):329-342.

33. Sugita Y (2008) Face perception in monkeys reared with no exposure to faces. *Proc Natl Acad Sci USA* 105(1):394-398.

34. Polk TA, Park J, Smith MR, Park DC (2007) Nature versus nurture in ventral visual cortex: a functional magnetic resonance imaging study of twins. *J Neurosci* 27(51):13921-13925.

35. Wilmer JB, et al. (2010) Human face recognition ability is specific and highly heritable. *Proc Natl Acad Sci USA* 107(11):5238-5241.

36. Le Grand R, Mondloch CJ, Maurer D, Brent HP (2004) Impairment in holistic face processing following early visual deprivation. *Psychol Sci* 15(11):762-768.

37. Boremanse A, Norcia AM, Rossion B (2013) An objective signature for visual binding of face parts in the human brain. *J Vis* 13(11):1-18.

38. Rossion B (2008) Picture-plane inversion leads to qualitative changes of face perception. *Acta Psychol* 128(2):274-289.

39. McKone E, Yovel G (2009) Why does picture-plane inversion sometimes dissociate perception of features and spacing in faces, and sometimes not? Toward a new theory of holistic processing. *Psychon. B. Rev.* 16(5):778-797.

40. Tan C, Poggio T (2013) Faces as a "Model Category" for Visual Object Recognition. *MIT-CSAIL-TR-2013-004 / CBCL-311*. URI: http://hdl.handle.net/1721.1/77936

**Figure Legends**

**Figure 1. Tuning size determines whether processing is "face-like" or "object-like", as gauged by the Composite Face Effect (CFE)**. **a,** Aligned composite faces. Top halves are identical, while bottom halves are different. People sometimes incorrectly perceive the two identical top halves as different. **b,** Misaligned composite faces. Human judgement of the top halves (as being identical) is significantly more accurate for misaligned than aligned composites (15, 16). This is the CFE, which is only found for faces. **c,** Simulations show that the CFE is produced by neurons with large – but not small – tuning size. Large: $p<0.001$, small: $p=0.85$. No CFE was found in either condition for inverted images (20); large: $p=0.39$, small: $p=0.81$. **Inset:** each individual neuron with large tuning size can produce the CFE by itself ($p<0.001$, both with and without controlling for inverted images). Error bars: SEM.

**Figure 2. Tuning size accounts for the behavioral Face Inversion Effect (FIE). a, b,** Illustration of the FIE: dissimilarity between faces is more apparent for upright than inverted faces (23, 25). **c,** Simulations show that decrease in dissimilarity (as a result of inversion) varies with tuning size. **d,** FIE effect size (upright dissimilarity – inverted dissimilarity) varies with tuning size. Large vs. medium: $p<0.0001$. Medium vs. small: $p<0.0001$. Neurons with small tuning size show "object-like" processing, i.e. a minimal inversion effect. **e**, Dissimilarity after controlling for coverage area of different tuning

sizes (see Methods). **f**, FIE effect size, controlling for coverage area. Large vs. medium: p=0.023. Medium vs. small: p=0.012. Large vs. small: p=0.0002. Error bars: SEM.

**Figure 3. Tuning size accounts for the neural Face Inversion Effect (FIE). a,** Mean individual neuron responses to single faces (as opposed to dissimilarities between pairs of faces, in Fig. 2) for upright vs. inverted faces. **b,** FIE effect size (upright response – inverted response) varies with tuning size. Large vs. medium: $p<0.0001$. Medium vs. small: $p<0.0001$. **c**, Mean individual neuron responses, after controlling for response in the upright condition (see Methods). **d**, FIE effect size, after controlling for response in the upright condition. Large vs. medium: $p<0.0001$. Medium vs. small: $p<0.0001$. Error bars: SEM.

**Figure 4. Tuning size accounts for the Whole-Part Effect (WPE). a**, Study face. **b**, Whole condition: two choices (differing only in the eye region) are presented, to be matched to the study face from memory. **c**, Part condition: two choices (eye regions cropped from Whole condition) to be matched to the study face. The WPE is defined as a significantly higher accuracy for the Whole condition than the Part condition (16). **d**, Simulation results. Large: $p<0.001$, small: $p<0.001$. **e**, The WPE effect size (Whole accuracy – Part accuracy) varies with tuning size. Large vs. small: $p<0.001$. This remains significant after subtracting effect sizes for inverted images (20); large vs. small: $p<0.001$. Error bars: SEM.

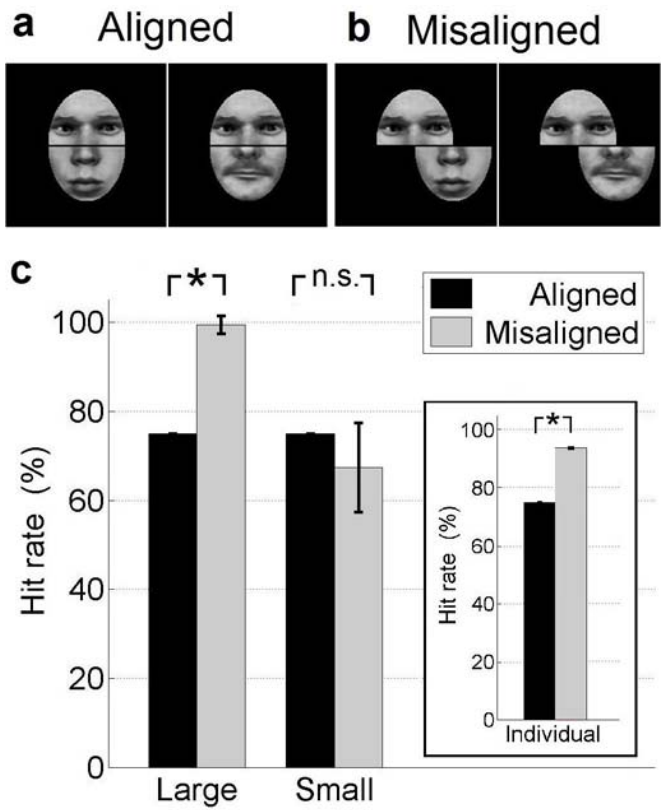

**Figure 1. Tuning size determines whether processing is "face-like" or "object-like", as gauged by the Composite Face Effect (CFE)**.

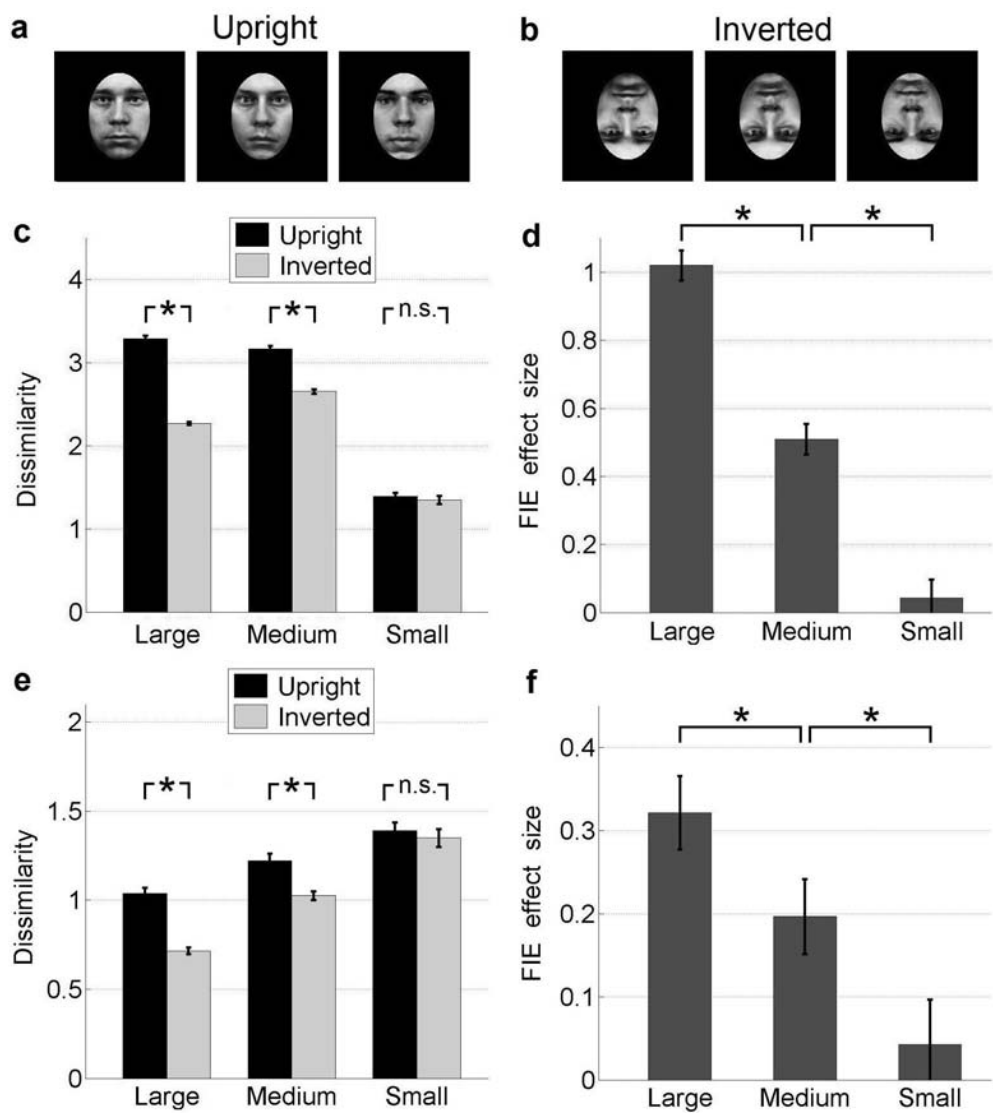

**Figure 2. Tuning size accounts for the behavioral Face Inversion Effect (FIE).**

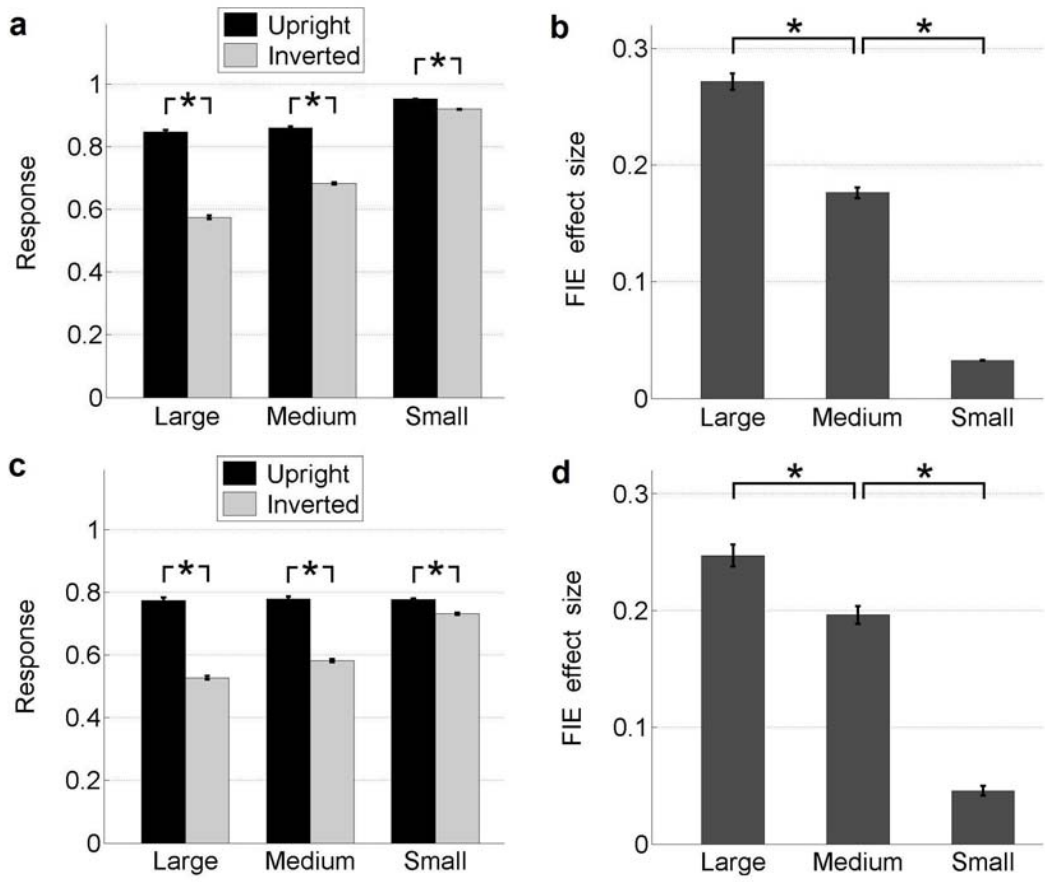

**Figure 3. Tuning size accounts for the neural Face Inversion Effect (FIE).**

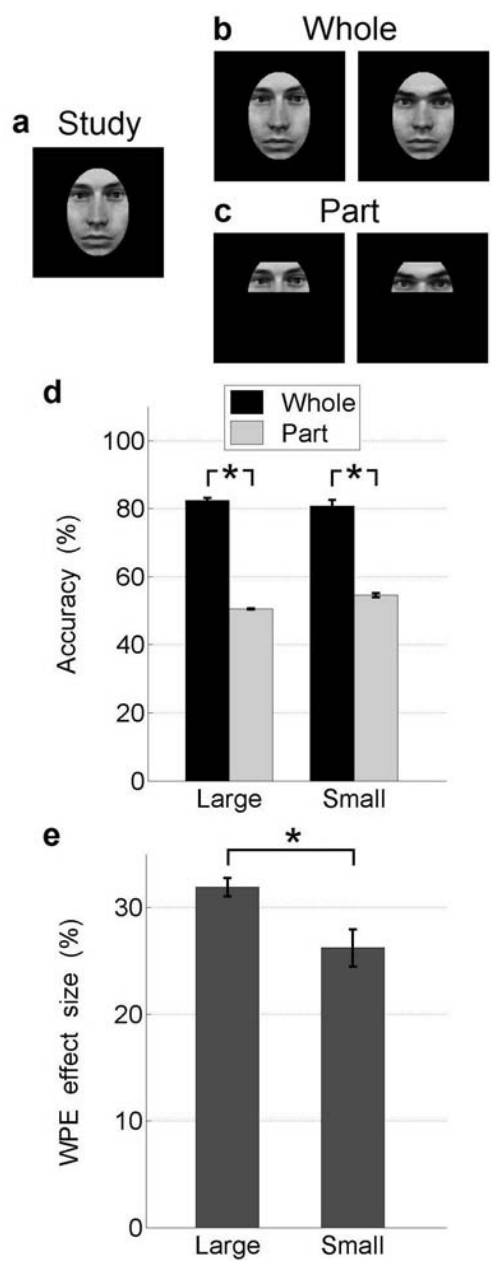

**Figure 4. Tuning size accounts for the Whole-Part Effect (WPE).**